# Explainable artificial intelligence model predicting the risk of all-cause mortality in patients with type 2 diabetes mellitus

Olga Vershinina[1,2*], Jacopo Sabbatinelli[3,4*], Anna Rita Bonfigli[5], Dalila Colombaretti[3], Angelica Giuliani[3], Mikhail Krivonosov[1,2], Arseniy Trukhanov[6], Claudio Franceschi[2], Mikhail Ivanchenko[1,2†], Fabiola Olivieri[3,7†]

[1] *Research Center in Artificial Intelligence, Institute of Information Technologies, Mathematics and Mechanics, Lobachevsky State University, Nizhny Novgorod 603022, Russia*
[2] *Institute of Biogerontology, Lobachevsky State University, Nizhny Novgorod 603022, Russia*
[3] *Department of Clinical and Molecular Sciences, DISCLIMO, Università Politecnica delle Marche, Ancona 60121, Italy*
[4] *Clinic of Laboratory and Precision Medicine, IRCCS INRCA, Ancona 60121, Italy*
[5] *Scientific Direction, IRCCS INRCA, Ancona 60121, Italy*
[6] *Mriya Life Institute, National Academy of Active Longevity, Moscow 124489, Russia*
[7] *Advanced Technology Center for Aging Research, IRCCS INRCA, Ancona 60121, Italy*

*\* These authors share first authorship*
*† These authors share senior authorship*

**Abstract**
**Objective.** Type 2 diabetes mellitus (T2DM) is a highly prevalent non-communicable chronic disease that substantially reduces life expectancy. Accurate estimation of all-cause mortality risk in T2DM patients is crucial for personalizing and optimizing treatment strategies.
**Research Design and Methods.** This study analyzed a cohort of 554 patients (aged 40–87 years) with diagnosed T2DM over a maximum follow-up period of 16.8 years, during which 202 patients (36%) died. Key survival-associated features were identified, and multiple machine learning (ML) models were trained and validated to predict all-cause mortality risk. To improve model interpretability, Shapley additive explanations (SHAP) was applied to the best-performing model.
**Results.** The extra survival trees (EST) model, incorporating ten key features, demonstrated the best predictive performance. The model achieved a C-statistic of 0.776, with the area under the receiver operating characteristic curve (AUC) values of 0.86, 0.80, 0.841, and 0.826 for 5-, 10-, 15-, and 16.8-year all-cause mortality predictions, respectively. The SHAP approach was employed to interpret the model's individual decision-making processes.
**Conclusions.** The developed model exhibited strong predictive performance for mortality risk assessment. Its clinically interpretable outputs enable potential bedside application, improving the identification of high-risk patients and supporting timely treatment optimization.

**Article Highlights**

- **Why did we undertake this study?**
- Accurate prediction of mortality risk in patients with type 2 diabetes mellitus (T2DM) is critical for optimizing personalized treatment approaches and improving clinical outcomes.
- **What is the specific questions we wanted to answer?**
  The primary objectives of this study were to identify clinically significant prognostic factors associated with survival in T2DM patients, and develop an explainable predictive model for mortality risk assessment.
- **What did we find?**
  Our analysis identified ten clinically significant predictors and yielded a high-performance predictive model for all-cause mortality risk assessment in T2DM patients.
- **What are the implications of our findings?**
  The explainable nature (SHAP analysis) and strong predictive performance, positions our model as a potentially valuable clinical decision-support tool to enhance T2DM patient management.

**Introduction**

Diabetes mellitus is a prevalent endocrine-metabolic disorder characterized by chronic hyperglycemia resulting from either impaired insulin secretion or insulin resistance. The global diabetes epidemic continues to escalate at an alarming rate, imposing substantial strain on healthcare systems worldwide. According to International Diabetes Federation estimates, the worldwide prevalence of DM reached 536.6 million cases in 2021, with projections indicating a dramatic rise to approximately 783.2 million cases by 2045 (1). Type 2 diabetes mellitus (T2DM) represents the most prevalent form of diabetes, comprising 90-95% of all diagnosed cases. This metabolic disorder is strongly correlated with obesity and physical inactivity, which are among its primary modifiable risk factors (2). Patients with T2DM exhibit significantly elevated risks of diabetes-related complications (3) and demonstrate higher all-cause and cause-specific mortality rates, particularly from cardiovascular disease, when compared to both the general population and non-diabetic individuals (4–6). Current evidence indicates that excess mortality in diabetic patients can be effectively reduced through optimal pharmacotherapy and lifestyle interventions (7,8). Accurate prediction of individual mortality risk in DM therefore serves as a critical foundation for developing personalized therapeutic approaches aimed at improving both life expectancy and quality of life.

Contemporary mortality risk prediction increasingly utilizes machine learning (ML) and artificial intelligence (AI) approaches. Current mortality risk assessment in T2DM patients predominantly employs Cox proportional hazards regression models (9–18). However, emerging approaches utilizing advanced ML algorithms (e.g., random forests, neural networks) offer distinct advantages by capturing non-linear risk function dependencies that Cox regression cannot accommodate (19–21). These ML-based models demonstrate robust predictive performance, with several achieving accurate long-term mortality risk stratification over 10-15 year horizons (15–17,19–21).

Nevertheless, current mortality prediction models suffer from limited transparency and interpretability, frequently functioning as "black boxes" that compromise clinical trust and impede practical implementation. While explainable AI (XAI) techniques have been incorporated into only two diabetes mortality prediction models to date (22,23) – encompassing both type 1 and type 2 diabetes – these implementations merely explain aggregate model behavior rather than providing patient-specific interpretations. To bridge this critical gap, we sought to develop a novel, interpretable AI framework capable of generating individualized explanations for long-term all-cause mortality risk predictions in T2DM patients.

**Research Design and Methods**
**Study Population**
The study sample was drawn from a previously established cohort comprising 568 patients diagnosed with T2DM (17,24). The patients were recruited at the Metabolic Diseases and Diabetology Department of IRCCS INRCA (Ancona, Italy) between May 2003 and November 2006. T2DM was diagnosed according to American Diabetes Association (ADA) criteria, which included any of the following: hemoglobin A1c (HbA1c) level ≥6.5% (48 mmol/mol), fasting blood glucose ≥126 mg/dL, 2-hour blood glucose ≥200 mg/dL during oral glucose tolerance test (OGTT), or random blood glucose ≥200 mg/dL in the presence of severe diabetes symptoms (2). Inclusion criteria for patients with diabetes were age from 40 to 87 years, a body mass index (BMI) <40 kg/m$^2$, ability and willingness to give written informed consent. The study was approved by the Institutional Review Board of IRCCS INRCA hospital (Approval no. 34/CdB/03) and conducted in accordance with the principles outlined in the Declaration of Helsinki.

**Outcomes**
All-cause mortality data were extracted from medical records spanning enrollment through December 31, 2019. Overall survival time was calculated from enrollment to death. For surviving patients, follow-up duration was censored at their last recorded observation. The maximum follow-up period was 16.8 years (6142 days).

**Covariates**
Baseline information collected at enrollment included clinical characteristics such as age, sex, anthropometric parameters, smoking, and medical history (duration of T2DM, presence of comorbidities and complications of diabetes, concurrent treatments). Comorbidities included arterial hypertension and dyslipidemia. Complications of diabetes were diabetic neuropathy, diabetic nephropathy, diabetic retinopathy, atherosclerotic vascular disease, and major adverse cardiovascular events (MACE). Fasting blood samples from all participants were processed to obtain serum and stored at −80 °C. All serum samples were screened for hemolysis prior to analysis. In all participants, standard methods were utilized to assess blood cell counts and biochemical parameters. Serum biomarkers were measured using standardized CE-IVD assays. The serum N-glycomic profile was assessed using a validated method based on IgG purification with protein G, enzymatic release of N-glycans by PNGase F, fluorescent labeling with 2-aminobenzamide (2-AB), and chromatographic separation, as previously described (25).

**Prediction Model Development**

The dataset was preprocessed before applying ML algorithms. Covariates (variables/features) with >20% missing values were removed, along with samples missing data for age, sex, disease duration, survival information, or categorical features. After filtering, 123 features and 554 patients remained. The dataset was split into training and testing sets at an 80:20 ratio through stratified random sampling based on survival status. Data were then z-normalized using means and standard deviations derived from the training set. Finally, remaining missing values were imputed using the k-nearest neighbors algorithm (k=5).

We performed feature selection on the training data to remove weakly predictive variables using four approaches: mutual information, spatially uniform reliefF, and minimum redundancy-maximum relevance (each retaining the top 50% of ranked features), plus univariate Cox regression (Benjamini-Hochberg-adjusted p-values <0.05). The final feature set combined the intersection of these methods' outputs, further refined through model-specific forward selection.

Nine ML algorithms were used to predict the risk of all-cause mortality: multivariate Cox proportional hazards model with ridge penalty (CoxPH), random survival forest (RSF), extra survival trees (EST), component-wise gradient boosting (CWGB), gradient-boosted regression trees (GBRT), extreme gradient boosting survival embeddings (XGBSE), and three artificial neural networks – Cox proportional hazards deep neural network (DeepSurv), case-control Cox regression model (CoxCC), and piecewise constant hazard model (PCHazard).

The hyperparameters of the models were tuned using the multivariate tree-structured Parzen estimator. The total number of optimization trials was 100. The best trial with the optimal combination of hyperparameters was defined using the 5-fold stratified cross-validation procedure on the training dataset. Model performance was evaluated using Harrell's concordance index (C-index) as the primary metric. We additionally conducted time-dependent receiver operating characteristic curve (ROC) analysis to calculate area under the curve (AUC) values and assessed calibration via the Integrated Brier Score (IBS).

We performed both global and local interpretability analysis of the optimal model using Shapley additive explanations (SHAP), with all surviving patients from the training dataset serving as the background distribution for SHAP value computation. All modeling workflows – including development, evaluation, and interpretation – were implemented in Python 3.11.7 and R 4.3.2.

**Statistical Analysis**

We compared survival groups (alive vs. deceased) using Mann-Whitney U tests for continuous variables and $\chi^2$ tests for categorical variables, with statistical significance set at Benjamini-Hochberg-adjusted p-values <0.05 (two-sided). Survival analysis between risk groups (stratified by median predicted risk scores from training data) employed Kaplan-Meier estimation and log-rank testing.

## Results
### Cohort Analysis
After preprocessing, the final dataset included 554 patients (302 male, 252 female) with a median age of 67 years (interquartile range, IQR 61-72) at baseline. The median T2DM duration was 14 years (IQR 7-21, range 1-54). During the 16.8-year follow-up, 202 deaths occurred (40 within 5 years, 94 within 10 years, 178 within 15 years), with deceased patients showing median survival of 10.6 years (IQR 6.3-13.6). Among 352 survivors, only 3 patients were lost to follow-up. Comparative analysis of 123 baseline characteristics revealed 36 statistically significant differences between surviving and deceased patients.

### Development of the Model for Predicting Mortality Risk
Feature selection identified 16 variables consistently ranked as important across all four methods. Notably, three features – age, N-terminal prohormone of brain natriuretic peptide (NT-proBNP), and high-sensitivity troponin I (hs-cTnI) – overlapped with a previously published Cox model from the same dataset (17). We additionally incorporated three prognostic factors from this model, hemoglobin A1c (HbA1c), C-reactive protein (hs-CRP), and soluble suppression of tumorigenicity 2 (sST2) (17), and obtained an intermediary set of 19 features for ML.

Then, for each of the nine ML models, we performed forward feature selection to identify the optimal subset from the 19 candidate features. The hyperparameters of each model were fine-tuned for every tested feature subset to maximize C-index. The evaluation results are summarized in Table 1. Among all models, the EST model demonstrated superior performance in both cross-validation and the test dataset. On the training data, cross-validation yielded a C-index of 0.751 and a 16.8-year AUC of 0.791. When evaluated on the test dataset, the EST model achieved a C-index of 0.776 and a 16.8-year AUC of 0.826, further confirming its robustness. Additionally, IBS of 0.1 indicates good calibration.

The optimal EST model was trained using ten key variables: age, number of complications, NT-proBNP, triglycerides, creatinine, hs-CRP, RDW-SD, apolipoprotein A1, disease duration, and the relative abundance of a specific serum N-glycan structure – NA3F, a triantennary, α-1,3 core-fucosylated, branched N-glycan derived from glycoproteins. Survival curves of the high-risk and low-risk groups are shown in Fig. 1A, B. In both the training and test datasets, overall survival is statistically significantly higher in the low-risk group.

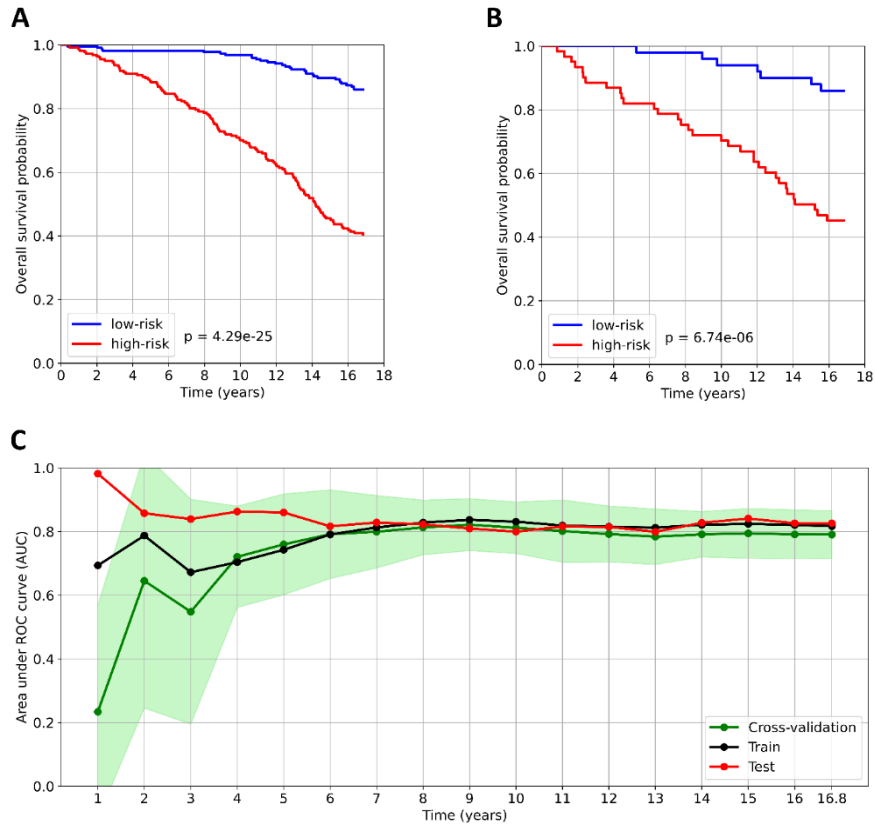

**Figure 1** – Analysis of the predictive model for all-cause mortality in patients with type 2 diabetes. A: and B: Kaplan–Meier survival curves for the low-risk and high-risk groups in the train and test datasets, respectively. Patients were stratified into risk groups based on the median predicted risk score derived from the training dataset. C: Time-dependent AUC over the observation period. The AUC values calculated obtained from cross-validation are presented as the mean (green dots) ± standard deviation (light green area).

The developed prediction model demonstrates robust performance for both medium- and long-term mortality risk predictions, with time-dependent AUC values consistently exceeding 0.8 for forecast periods beyond five years (Fig. 1C). Specifically, the test dataset achieved AUCs of 0.86, 0.80, and 0.84 at 5, 10, and 15 years, respectively. However, for time intervals shorter than five years, we observed a notable discrepancy between the test dataset AUC and those derived from both the training dataset and cross-validation. This discrepancy stems from two key factors. First, the model was explicitly optimized for 16.8-year mortality risk prediction, resulting in reduced reliability for short-term forecasts. Second, the dataset contained only 40 patients who died within the first five years of follow-up, leading to overly limited training data and potential bias in early-term predictions.

**Interpretation of a Model Predicting Mortality Risk**

We analyzed SHAP values to interpret the contribution of the ten selected features in predicting 16.8-year mortality risk among patients with T2DM. This global explainability analysis of the EST model quantified the relative importance of each feature in the model's predictions. Fig. 2A presents the mean absolute SHAP values, representing the average contribution magnitude of each feature to the model's predictions. Age, number of complications, and disease duration emerged as the strongest predictors of mortality risk, followed by laboratory biomarkers. Fig. 2B illustrates the directional effects of these features, where positive SHAP values correspond to increased mortality risk and negative values indicate protective effects. Notably, apolipoprotein A1 showed an inverse relationship with 16.8-year mortality risk, where elevated levels were associated with reduced mortality probability. The remaining nine features demonstrate positive associations with predicted mortality risk. However, it should be emphasized that while SHAP analysis reveals these important feature-prediction relationships, it does not imply causation – it only identifies associations between variables and model outputs.

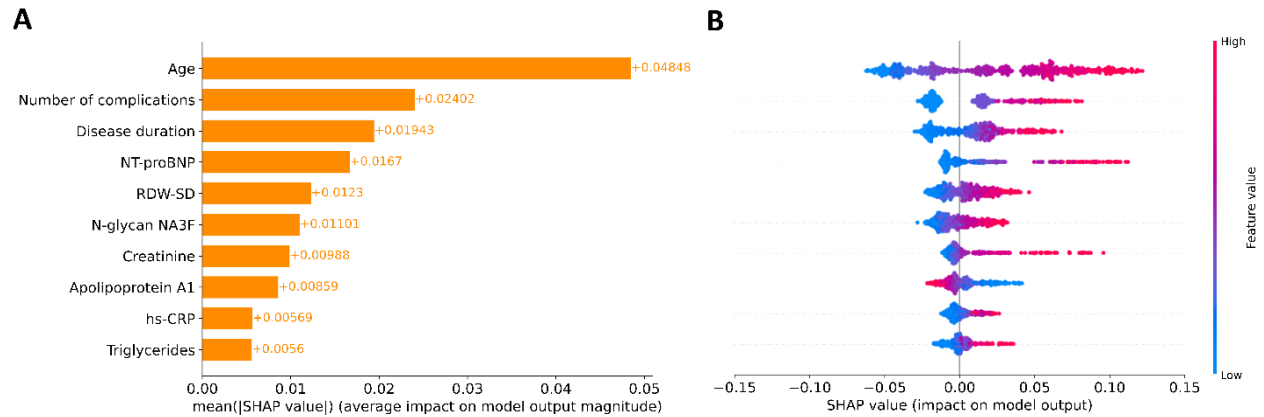

**Figure 2** – Global explanation of feature contributions to model predictions. A: Feature importance ranking based on mean absolute SHAP values across all participants. Features are ordered vertically by their relative impact on model predictions, with the most influential at the top. B: SHAP summary plot showing the directional relationship between feature values and model outputs. Individual points represent SHAP values for each feature-participant combination, with color intensity indicating feature values (red: high, blue: low).

For each feature, we determined thresholds at which SHAP values change sign. In individuals older than 64, the predicted probability of mortality increased. An increase in complication number was associated with an increased probability of mortality. The mortality risk increases when patients have had T2DM for more than 9 years. As for laboratory parameters, the values contributing to an increased risk of mortality include: NT-proBNP >100 ng/L, RDW-SD >42.3 fL, N-glycanNA3F >2.7%, creatinine >1.0 mg/dL, apolipoprotein A1 <160 mg/dL, hs-CRP > 3.6 mg/L, and triglyceride >120 mg/dL.

SHAP values were also used to explain the model's decision-making process for individual predictions. Fig. 3A displays the local explainability plot for a long-term survivor (alive after 16.8 years) with a favorable predicted mortality probability (20.9%). All ten clinical factors contributed to risk reduction, with most important influences being relatively younger age, absence of diabetic complications, and low RDW-SD and creatinine values. The mirror image emerges in Fig. 3B, which explains the prediction for a deceased patient (death occurring 4.4 years post-examination), with a concerning 68.8% mortality risk. All features contributed to an increased risk, the strongest risk drivers were levels of creatinine and NT-proBNP, advanced age, and the burden of four diabetes-related complications.

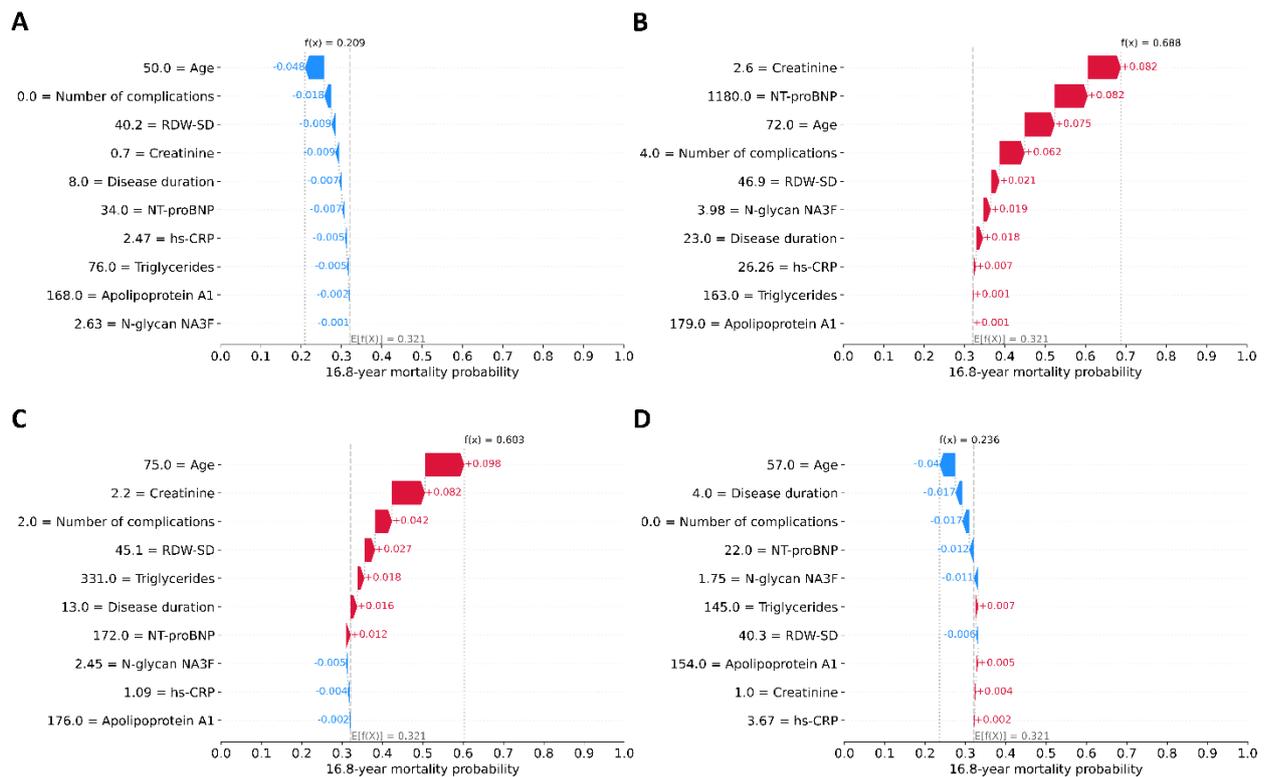

**Figure 3** – Local explanation of individual predictions using SHAP waterfall plots. Four representative cases are shown. A: a survivor (alive after 16.8 years) with low predicted mortality risk (20.9%), B: a deceased patient (death occurring 4.4 years post-examination) with high predicted risk (68.8%), C: a survivor with high predicted risk (60.3%), D: a deceased patient (death occurring 10.6 years post-examination) with low predicted risk (23.6%). The y-axis displays features ranked by their increasing predictive influence from bottom to top. Feature values are presented in their original scale for interpretability, though the model utilized normalized values internally. The x-axis represents the 16.8-year mortality probability. The prediction originates from the baseline probability E[f(X)] derived from the training set and subsequently modifies based on each feature's contribution. Each colored bar illustrates a feature's directional effect on the model's output: blue bars signify protective (risk-reducing) factors, while red bars denote hazardous (risk-increasing) factors.

Quality metrics demonstrate that our model exhibits strong predictive performance, consistently assigning lower risk scores to survivors and higher risk scores to deceased patients. However, certain cases may show significant prediction errors (either overestimation or underestimation of risk). In these instances, local explanation methods prove valuable for identifying the specific features responsible for these discrepancies. Fig. 3C displays the SHAP waterfall plot for a survivor with an unexpectedly high predicted mortality probability (60.3%). The analysis reveals that the elevated risk prediction was primarily driven by advanced age, elevated creatinine levels, presence of two diabetes-related complications, increased RDW-SD and triglyceride values, prolonged disease duration, and higher NT-proBNP concentration. In turn, Fig. 3D illustrates the SHAP analysis for a deceased patient (death occurring 10.6 years post-examination) where the model had predicted a low mortality probability (23.6%). The following factors contributed to this underestimation: younger age, shorter disease duration, lack of complications, lower levels of NT-proBNP, N-glycan NA3F, and RDW-SD.

While the precise reasons for these discrepancies remain unclear due to limited patient data, several potential explanations exist. The extended 16.8-year prediction window following baseline measurements introduces numerous unaccounted variables that could influence outcomes, including development of new complications or comorbidities, changes in treatment adherence, lifestyle modifications, and other unreported clinical factors. Nevertheless, the model's strong performance in long-term predictions suggests these confounding factors have relatively modest effects overall.

**Conclusions**

In this study, we developed an explainable AI model using the extra survival trees algorithm to predict 16.8-year all-cause mortality risk in patients with T2DM. In the test dataset, our model demonstrated strong predictive performance across all time horizons, with AUC values of 0.86 (5-year), 0.80 (10-year), 0.841 (15-year), and 0.826 (16.8-year) for all-cause mortality prediction. The overall concordance index (C-index) reached 0.776, with excellent calibration (IBS = 0.1). Notably, this represents a significant improvement over the previously developed Cox regression-based nomogram when evaluated on the same dataset (17).

The final model variables incorporated age, number of complications, disease duration, NT-proBNP, RDW-SD, N-glycan NA3F, creatinine, apolipoprotein A1, hs-CRP, and triglycerides. These variables have been previously employed in various combinations across 15 existing mortality prediction studies (9–23). Age consistently appeared in all 15 models. While no studies directly included complication count as a variable, several incorporated specific complications (10,11,15,18,19,22). Among other predictors, triglycerides featured in seven models (10,12,15,19–21,23), diabetes duration in five (13–15,22,23), creatinine in three (16,21,23), hs-CRP in two (17,21), and NT-proBNP in one model (17). Notably, sex – which was selected in all studies except one (12), did not in our study.

Our model combines ML-driven accuracy with SHAP-based interpretability, revealing both global feature importance and directional effects on 16.8-year mortality risk. While elevated apolipoprotein A1 decreases predicted risk, the other nine features (e.g., age, creatinine) show positive associations – all consistent with established T2DM mortality relationships.

In older people with diabetes, additional factors such as increased diabetes complications, polypharmacy, physical and mental frailty are present, contributing to an increase in the number of deaths (26–28). A higher number of diabetes-related complications significantly correlates with increased mortality risk (29). Similarly, the risk of all-cause mortality and cardiovascular disease mortality significantly increases with T2DM duration (30).

The model seems to capture broader dimensions of biological aging and residual risk, integrating emerging concepts such as residual inflammatory risk (RIR) (31) and organ-specific ageotyping (32). Elevated hs-CRP levels, which contributed to increased mortality risk in our model, are consistent with the notion of RIR and its clinical relevance in cardiovascular prevention. The SHAP-derived threshold is consistent with previously proposed cut-offs for cardiovascular risk (33), reinforcing the role of low-grade inflammation as a relevant prognostic factor in T2DM.

NT-proBNP, a validated cardiac biomarker, captured the contribution of subclinical myocardial stress in our model. Beyond its role in diagnostics and management of heart failure, elevated levels also reflect chronic hemodynamic strain and myocardial remodeling, indicating cardiac aging, and aligning with our broader hypothesis that progressive cardiac dysfunction may represent an expression of biological aging mechanisms in T2DM (34).

Creatinine, a conventional marker of renal function, may serve as a proxy for biological aging of the kidney. While glomerular filtration rate physiologically declines with age, patients with T2DM experience an accelerated reduction, reflecting premature renal dysfunction (35). This renal trajectory often parallels that of the heart, as the interplay between cardiac and renal aging is well established and clinically recognized in the context of cardiorenal syndromes (36).

Red cell distribution width-standard deviation (RDW-SD), a measure of anisocytosis, also emerged as a relevant predictor. Although traditionally used in the evaluation of anemia, elevated RDW has been associated with cardiovascular events and mortality (37). Chronically elevated RDW is increasingly regarded as a marker of bone marrow stress, potentially reflecting impaired erythropoiesis in the setting of chronic inflammation and immune activation. In this context, it may represent a hematopoietic expression of biological aging.

Lipid-related biomarkers such as triglycerides and apolipoprotein A1 (ApoA1) were also retained in the model and showed opposing associations with mortality risk. Elevated triglyceride levels are a hallmark of insulin resistance and atherogenic dyslipidemia, and their association with cardiovascular and all-cause mortality has been consistently observed in patients with T2DM (38). ApoA1, the main apolipoprotein component of high-density lipoprotein (HDL) particles, was inversely associated with mortality risk in our model. Reduced circulating levels of ApoA1 have been associated with increased risk of incident diabetes (39) as well as with cardiovascular events in large general population cohorts (40), although its prognostic value has not been clearly demonstrated in diabetic populations. In this context, both triglycerides and ApoA1 may act as complementary indicators of residual lipid-related risk, particularly relevant in patients receiving statin therapy, as was the case for the vast majority of our cohort.

N-glycan NA3F was associated with metabolic and inflammatory features in T2DM (25). Although the biological role of this structure remains elusive, its inclusion may reflect broader N-glycan remodeling processes linked to aging, immune regulation, or glycoprotein turnover, underscoring the potential of serum glycomics to capture latent biological signals beyond conventional biomarkers.

Limitations of our study include a moderate sample size, its origin from a single medical center, and Italian ancestry of patients might restrict the generalizability of the findings. Information regarding the specific causes of death and some potential predictors, such as diet and physical exercise was not available. Although the model's performance was assessed through both cross-validation and an independent internal testing dataset, a suitable external validation dataset was unavailable. These limitations should be considered when interpreting the results of our study.

In conclusion, this study presents a novel the ML model that predicts the risk of 16.8-year all-cause mortality in patients with T2DM, utilizing ten clinical and laboratory parameters. Taken together, the model variables reflect a multidimensional construct of long-term risk in T2DM, incorporating diverse but interconnected processes related to biological aging, residual inflammation, and subclinical organ dysfunction. Their influence on individual patient predictions is disclosed by the local explanation SHAP method, which has not been done previously in existing all-cause mortality prediction models for patients with T2DM. Thus, our explainable model can be potentially used as an additional tool in the examination of patients with T2DM.

**Acknowledgements**
**Funding.** This work was supported by the Ministry of Economic Development of the Russian Federation (grant No 139-15-2025-004 dated 17.04.2025, agreement identifier 000000C313925P3X0002).
**Duality of Interest.** No potential conflicts of interest relevant to this article were reported.
**Author Contributions.** O.V., J.S., C.F., M.I., and F.O. were involved in the conception, design, and methodology of the study and the interpretation of the results. A.T., C.F., M.I., and F.O. supervised the study. J.S., A.R.B., D.C., A.G., and F.O. contributed to the data curation. O.V., J.S., and M.K. performed the data analysis. O.V. developed the predictive model and software. O.V. and J.S. wrote the first draft of the manuscript. All authors edited, reviewed, and approved the final version of the manuscript. F.O. is the guarantor of this work and, as such, has full access to all the data in the study and takes responsibility for the integrity of the data and the accuracy of the data analysis.

**Tables**

**Table 1** – C-index scores of ML models predicting all-cause mortality in patients with type 2 diabetes

| Model | Number of selected features | C-index, cross-validation | C-index, train | C-index, test |
| --- | --- | --- | --- | --- |
| EST | 10 | 0.7511 | 0.7697 | 0.7763 |
| DeepSurv | 13 | 0.7509 | 0.7676 | 0.7638 |
| CoxCC | 12 | 0.7486 | 0.7631 | 0.7449 |
| CoxPH | 10 | 0.7485 | 0.7515 | 0.7468 |
| CWGB | 12 | 0.7451 | 0.7509 | 0.7407 |
| RSF | 8 | 0.7441 | 0.7636 | 0.7415 |
| PCHazard | 19 | 0.7428 | 0.7705 | 0.6814 |
| XGBSE | 10 | 0.7424 | 0.7903 | 0.7369 |
| GBRT | 11 | 0.7367 | 0.8283 | 0.7205 |

Models are ranked in descending order of the C-index score calculated using cross-validation. EST, extra survival trees; DeepSurv, Cox proportional hazards deep neural network; CoxCC, case-control Cox regression model; CoxPH, multivariate Cox proportional hazards model with ridge penalty; CWGB, component-wise gradient boosting; RSF, random survival forest; PCHazard, piecewise constant hazard model; XGBSE, extreme gradient boosting survival embeddings; GBRT, gradient-boosted regression trees.